# Generative AI Application for Building Industry

Hanlong Wan*, Jian Zhang, Yan Chen, Weili Xu, Fan Feng

Pacific Northwest National Laboratory, Richland, WA, USA
*Corresponding Author
hanlong.wan@pnnl.gov

Notice: This material is based on work supported by Battelle Memorial Institute under Contract No DE-AC05—76RL01830 with the U.S. Department of Energy. The U.S. Government retains, and the publisher by accepting this article for publication, acknowledges that the U.S. Government retains a non-exclusive, paid-up, irrevocable, world-wide license to publish or reproduce the published form of this manuscript, or allow others to do so, for U.S. Government purposes.

**ABSTRACT** (200-250 words)

This paper investigates the transformative potential of generative AI technologies, particularly large language models (LLMs), within the building industry. By leveraging these advanced AI tools, the study explores their application across key areas such as energy code compliance, building design optimization, and workforce training. The research highlights how LLMs can automate labor-intensive processes, significantly improving efficiency, accuracy, and safety in building practices. The paper also addresses the challenges associated with interpreting complex visual and textual data in architectural plans and regulatory codes, proposing innovative solutions to enhance AI-driven compliance checking and design processes. Additionally, the study considers the broader implications of AI integration, including the development of AI-powered tools for comprehensive code compliance across various regulatory domains and the potential for AI to revolutionize workforce training through realistic simulations. This paper provides a comprehensive analysis of the current capabilities of generative AI in the building industry while outlining future directions for research and development, aiming to pave the way for smarter, more sustainable, and responsive construction practices.

# NOMENCLATURE[†]

| | |
|---|---|
| AEC | Architecture, Engineering, and Construction |
| AI | Artificial Intelligence |
| BEM | Building Energy Modeling |
| BIM | Building Information Modeling |
| DOE | Department of Energy |
| DQN | Deep Q-Network |
| GAN | Generative Adversarial Network |
| HVAC | Heating, Ventilation, and Air Conditioning |
| IDF | Input Data File |
| LLM | Large Language Model |
| NLP | Natural Language Processing |
| OCR | Optical Character Recognition |
| PCS | Physical and Computational Sciences |
| RAG | Retrieval-Augmented Generation |
| UBEM | Urban Building Energy Modeling |
| VAE | Variational Autoencoder |

---

[†] The full names of models, such as ChatGPT (Chat Generative Pre-Trained Transformer), are not included in this list as they are already widely recognized by the public.

# 1 INTRODUCTION

U.S. buildings, which consume a substantial 38% of the nation's total energy usage [1], present a significant opportunity for decarbonization and efficiency improvements. Artificial Intelligence (AI) stands as a pivotal tool in transforming the building sector. This underscores the urgency and potential for integrating Generative AI, refers to computational techniques that are capable of generating seemingly new, meaningful content [2], across building systems to catalyze advancements in energy management [3].

Generative AI has been successfully applied in areas like Code Copilot [4], translation [5], and game computer graphics [6], due to its ability to generate context-aware, high-quality outputs with minimal human input. Compared to traditional methods, it excels in automating complex tasks, improving accuracy, and significantly enhancing both speed and scalability in processes like coding, language translation, and graphical design. Thus, Generative AI can also play a critical role in building industry, including building design [7], construction [8], and operation [9]. The intricate mesh of design specifications, graphical data, and regulatory requirements provides a fertile ground for Generative AI applications that can streamline processes, optimize material use, and ensure compliance with stringent safety and construction standards. However, these challenges include the availability of robust datasets to train AI models, the accuracy and sensitivity of AI systems, and the alignment of these technologies with evolving energy efficiency assessments [10,11]

The present study aims to thoroughly examine the existing applications of Generative AI in the building industry, identifying both the strides and gaps in current research. This review will define pertinent research topics that align with Department of Energy (DOE) Building Technology Office (BTO) objectives for decarbonization and electrification [12] and the content prescribed by the President's Executive Order on the Safe, Secure, and Trustworthy Development and Use of AI [13]. Our goal is to formulate a strategic plan that leverages advanced Generative AI to address critical challenges in the building sector, ultimately enhancing sustainability and operational efficiency. This paper will pave the way for a transformative approach in building management, setting the stage for a future where AI-powered solutions drive the forefront of energy innovation, accuracy, and safety. Figure 1 shows the outline of this paper.

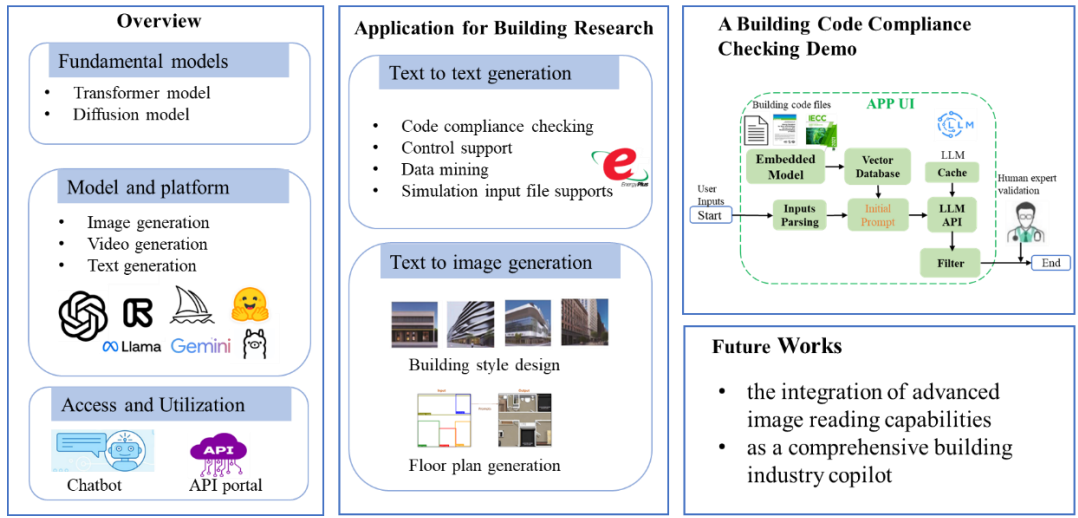

**Figure 1 Outline of the Manuscript**

The novelty of this work lies in its comprehensive analysis of the application of Generative AI across multiple domains within the building industry, offering a holistic view that extends beyond traditional studies focused on individual aspects of AI integration. By systematically examining how these advanced AI models can automate labor-intensive tasks such as energy code compliance checking, optimize design processes, and enhance workforce training through realistic simulations, this paper provides critical insights into the efficiency, accuracy, and safety improvements that AI can bring to building practices. Moreover, it highlights future directions for AI integration, emphasizing the development of tools that address current needs while anticipating future challenges in regulatory compliance and construction standards. As part of this work, we also developed a demonstration chatbot for building energy code compliance checking, which is applied to case studies to illustrate the potential of AI-driven solutions in the industry.

## 2 OVERVIEW OF CURRENT GENERATIVE AI

### 2.1 Fundamental Models

Diffusion model and transformer model are fundamental models in the context of Generative AI. They are among the core architectures that have significantly advanced the field. While they are not the only fundamental models used (others include General Adversarial Networks (GANs) [14], Variational Autoencoders (VAEs) [15], etc.), they are also pivotal in many cutting-edge applications today and engaging readers can read reference papers for more details.

#### 2.1.1 Transformer Model

Transformers are a type of neural network architecture that differ significantly from earlier recurrent neural architectures like long short-term memory (LSTM). Unlike LSTMs that process data sequentially, transformers process entire sequences of data simultaneously [16]. This parallel processing capability reduces training time, making transformers particularly efficient for handling large datasets. This efficiency has led to their widespread adoption in training models on extensive text corpora such as the Wikipedia database and the Common Crawl dataset [17]. Developing a machine learning model, including those based on transformers, is typically an iterative process where a data scientist might experiment with dozens to hundreds of model versions to achieve a model that satisfies predefined criteria [18].

A **Large Language Model (LLM)** is a type of Generative AI model, built upon transformer architecture, specifically designed to handle a variety of natural language processing (NLP) tasks, including text generation, language understanding, and classification [19], like GPT-4o [20]. As a widely-used example of Generative AI, LLMs are known for their ability to generate human-like text, making them integral to applications such as chatbots, text-to-image tools, and other AI-driven content creation technologies [21]. While LLMs offer promising enhancements in efficiency and decision-making within the building sector, their deployment must be carefully managed to address challenges related to data quality, security, and domain-specific requirements [22].

A Small Language Model (SLM) is a more resource-efficient alternative to LLMs, designed with significantly fewer parameters. SLMs excel in efficiency, requiring less computational power and producing a lower carbon footprint, making them more accessible to researchers and developers with limited resources [23]. However, they tend to underperform in complex tasks compared to LLMs, which can process longer sequences and capture intricate contextual relationships more effectively [23]. While SLMs offer a balance between performance and efficiency, they may fall short in scenarios requiring deep language understanding or few-shot learning.

### 2.1.2 Diffusion model

Diffusion models are a class of generative models that create data, such as images or text, by simulating a process that gradually transforms noise into meaningful content [24], like Stable Diffusion and DALL-E [25]. They work by first adding noise to the data in a forward process and then learning to reverse this process, effectively denoising the data to generate new, realistic samples. As a fundamental framework in Generative AI, diffusion models operate via this two-step process, where the forward diffusion transforms data into noise and the reverse process, guided by a neural network, reconstructs the data from noise [26]. These models have gained prominence in generating high-quality images by refining random noise into detailed and coherent outputs, surpassing traditional models like GANs and VAEs in stability and quality. However, diffusion models often require longer sampling times due to their iterative denoising process, making them computationally intensive [26]. Despite this, their ability to produce high-quality results across multiple domains, including image synthesis, text generation, and audio processing, makes them a powerful and versatile tool in generative AI.

## 2.2 Current Models and Platforms

In recent years, significant advancements in AI have given rise to highly sophisticated models capable of generating images, videos, and text. These models, primarily developed by leading IT companies, leverage the latest in diffusion and transformer technologies to produce remarkable results. This section provides an overview of the most notable models in these domains, their availability, and the various ways they can be utilized. Table 1 sum part of the models and platforms[‡].

**Table 1 Some Current Available Models and Platforms.**

| Company | Image Generation Model | Video Generation Model | Text Generation Model | Platform |
|---|---|---|---|---|
| OpenAI | DALL·E | Sora | GPT | OpenAI API Portal |
| Google | Imagen, Palm | Veo | Gemini* | Google DeepMind |
| Microsoft | NUWA-Infinity | NUWA-XL | Turing-NLG | Azure |
| Meta | Make-A-Scene | N/A | LLaMA* | N/A |
| Nvidia | Edify | N/A | Megatron-Turing NLG | Nvidia Playground |
| Tencent | GauGAN | N/A | Hunyuan | N/A |
| Alibaba | M6 | N/A | AliceMind, Qwen* | Alibaba Cloud |
| Baidu | ERNIE-ViLG | N/A | ERNIE | Baidu Cloud |
| Others | Stable Diffusion, Midjourney, Runway, etc. | Runway, Pixverse, Phenaki, MagicVideo, etc. | Anthropic Claude, etc. | Hugging Face*, Amazon SageMaker etc. |

*Open-source models

### 2.2.1 Image Generation Models

Image generation models have seen substantial improvements, with models like OpenAI's DALL·E 2 [27], Google's Imagen 3 [28], and Microsoft's NUWA-Infinity [29] leading the pack. DALL·E 2, for example, can create highly detailed and realistic images from textual descriptions, making it a powerful tool for artists and designers. Similarly, Google's Imagen 3 excels in generating photorealistic images with fine details and complex compositions. These models are available through online playgrounds

---

[‡] The information in this table was collected in July 2024. As website addresses may vary or be subject to change, the links are not all listed here. Readers interested in further details are encouraged to search the relevant terms using a search engine.

where users can experiment with generating images by inputting descriptive text. Additionally, APIs [30] provided by these companies allow developers to integrate image generation capabilities into their applications.

### 2.2.2 Video Generation Models

Video generation models, although still emerging, are showing great potential. Google's Veo [31] and Microsoft's NUWA-XL [29] are at the forefront, capable of creating high-quality videos from text prompts. These models can be used to generate content for entertainment, education, and marketing purposes. Like image generation models, video generation models are accessible through APIs, enabling seamless integration into various digital platforms. Additionally, platforms like Runway [32] offer user-friendly interfaces where creators can experiment with video generation without needing deep technical knowledge.

### 2.2.3 Text Generation Models

Text generation models have advanced significantly, with prominent examples including OpenAI's GPT-4 [33], Google's Gemini [34], and Meta's LLaMA [35]. These models leverage transformer architectures to produce coherent and contextually relevant text, making them invaluable for various applications such as chatbots, content creation, and more. Notably, models like LLaMA and Google's Gemini are available as open-source projects, allowing users to download and run them locally, providing flexibility for a wide range of use cases.

One of the most significant advancements in text generation is the adoption of **Retrieval-Augmented Generation** (RAG), a technique that greatly enhances the capabilities of language models by integrating external knowledge sources during the generation process [36]. Unlike traditional text Generation models, which rely solely on the pre-trained knowledge encoded within the model, RAG retrieves relevant information from indexed content to supplement the model's responses [37]. This approach allows for the generation of text that is not only more accurate but also more contextually relevant, as it incorporates up-to-date or domain-specific information that the model may not have encountered during its initial training.

RAG is versatile and can be applied to both SLMs and LLMs. By enabling these models to dynamically retrieve and utilize external data, RAG significantly improves their performance, making them particularly valuable in scenarios where the context, accuracy, and timeliness of information are critical [37].

The implementation of RAG can vary in complexity, as illustrated by different pipeline configurations in Table 2 [38]:
- Naive RAG involves basic indexing, standard retrieval methods based on query-document similarity, and straightforward text generation from the retrieved chunks.
- Advanced RAG introduces more sophisticated techniques, such as hierarchical indexing for more efficient data organization, query rewriting to improve retrieval precision, and the use of dense retrievers for better semantic understanding. It also includes post-retrieval processes like reranking and filtering to prioritize key information, ultimately leading to higher-quality text generation.
- Modular RAG represents a highly customizable approach, incorporating advanced indexing techniques like chunk optimization and hierarchical document organization. This configuration also supports pre-retrieval processes such as query expansion and transformation, and post-retrieval processes like compression and selection to refine the retrieved content. Additionally, Modular RAG includes options for fine-tuning retrievers and generators, routing queries through specific RAG pipelines, and using knowledge graphs to guide retrieval for enhanced responses.

**Table 2 Comparison between three RAG paradigms [38]**

| Pipeline | Module | Sub-Module/Process | Description |
|---|---|---|---|
| **Naive RAG** | Indexing | Basic Indexing | Organizes data chunks into a retrievable form. |
| | Retrieval | Standard Retrieval | Retrieves documents based on query-document similarity. |
| | Generation | Basic Generation | Generates responses based on retrieved chunks. |
| **Advanced RAG** | Indexing | Hierarchical Indexing | Structures information for efficient retrieval. |
| | Pre-Retrieval | Query Rewrite | Rewrites queries to improve retrieval precision. |
| | | HyDE | Hypothetical document generation to enhance retrieval accuracy. |
| | Retrieval | Dense Retriever | Retrieves documents using dense vectors for better semantics. |
| | Post-Retrieval | Rerank and Filter | Prioritizes key information, filtering out noise. |
| | Generation | Advanced Generation | Utilizes reranked chunks to generate high-quality responses. |
| **Modular RAG** | Indexing | Chunk Optimization | Refines chunk size and overlap to balance context and noise. |
| | | Structure Organization | Organizes documents hierarchically for better retrieval. |
| | Pre-Retrieval | Query Expansion | Expands queries to cover more retrieval contexts. |
| | | Query Transformation | Transforms original queries for more accurate results. |
| | Retrieval | Retriever FT | Fine-tuning retrievers for task-specific performance. |
| | | Retriever Source | Sources of retrieval include dense/sparse vectors and graphs. |
| | | Retriever Selection | Selects between dense, sparse, and hybrid retrievers. |
| | Post-Retrieval | Compression | Compresses retrieved chunks to retain key information. |
| | | Selection | Directly removes irrelevant chunks. |
| | Generation | Generator FT | Fine-tunes generators for task-specific output quality. |
| | | Verification | Verifies generated responses using external knowledge or models. |
| | Routing | Pipeline Routing | Routes queries to specific RAG pipelines based on content. |
| | Scheduling | Recursive or Iterative Retrieval | Schedules repeated retrieval/generation cycles until completion. |
| | Knowledge Guide | Knowledge Graph-Driven Reasoning | Guides retrieval using knowledge graphs for enhanced answers. |

**Content indexing** plays a crucial role in the effectiveness of both LLMs and smaller models [39]. Indexing organizes data into a retrievable structure, which is essential for RAG to function effectively [40]. Tools like LLaMA-index and other indexing frameworks allow for the efficient organization of content, enabling models to quickly access the relevant data needed for text generation. This process of indexing is fundamental, as it directly influences how well a model can retrieve and apply information during generation, thereby enhancing the overall quality and relevance of the generated content.

Supporting this ecosystem, online libraries such as Ollama [41] provide access to a range of models and tools, facilitating experimentation and development in AI-driven text generation. These resources make it easier for developers to implement and refine RAG techniques, ensuring that their models can deliver high-quality, contextually appropriate outputs.

In summary, the integration of RAG with advanced text generation models like GPT-4, combined with robust content indexing, represents a significant leap forward in the capabilities of AI-driven text generation. By overcoming the limitations of static training data, RAG enables LLMs and smaller models alike to generate more accurate, relevant, and up-to-date content, paving the way for smarter, more responsive AI applications.

### 2.3 Access and Utilization

There are several ways to access and utilize these advanced AI models, which can be broadly grouped into open and non-open solutions. Open-source models, such as Meta's LLaMA and Google's upcoming Gemini 2, can be downloaded and deployed locally, offering users full control over their AI capabilities. On the other hand, non-open platforms like Hugging Face and Azure Portal provide hosted environments where users can interact with models without needing to manage the underlying infrastructure.

Utilization can also be categorized based on the application type (or user-interface type). For instance, AI models can be integrated into chatbot development, as seen with GPT-based chatbots, or used within web portals like the Azure Portal, which allows for real-time interaction and model testing. Additionally, tools like GitHub Copilot leverage AI for code generation and assistance, while API access, such as through the OpenAI API, enables developers to seamlessly incorporate AI functionalities into their applications and workflows. These diverse options allow users to choose the best approach for their specific needs, whether they prioritize control, ease of use, or integration capabilities.

### 2.4 Common techniques for customizing models

In the field of generative AI, particularly in the deployment and optimization of LLMs, several techniques stand out: prompt engineering, RAG and embedding, fine-tuning, and continued pre-training (CPT), arranged in order of increasing complexity, quality, cost, and time. All play crucial roles in enhancing the performance and applicability of AI models across various domains and tasks.

**Prompt Engineering**

Prompt engineering refers to the process of designing, crafting, and refining inputs or prompts to optimize the responses of generative AI models, which are LLMs based on NLP [42]. The quality and structure of prompts are critical in guiding these models to produce meaningful, accurate, and contextually relevant outputs. In the realm of generative AI, especially in educational applications, prompt engineering becomes an essential skill that enables effective communication between humans and AI.

Generative AI's capabilities are vast, mimicking human language and comprehension, but the effectiveness of its responses largely depends on the quality of the prompts it receives. Well-crafted prompts can unlock the full potential of AI models, while poorly designed ones may lead to irrelevant or nonsensical outputs. This makes prompt engineering crucial, as it influences how the AI interprets tasks, ensures clarity in communication, and helps guide the AI toward desired outcomes [43]

In practical terms, prompt engineering has a direct impact on generative AI performance. By refining prompts to be clear, concise, and contextually rich, the AI can better understand the task and generate content that aligns with the user's needs. This optimization can be iterative, where prompts are tested, debugged, and fine-tuned to yield the most accurate and useful results. In educational contexts, for instance, prompt engineering helps in co-creation of content with AI, fostering deeper learning and more personalized interactions between AI and users [43].

**Retrieval System and Embedded Models**
Embedded models refer to the practice of integrating AI models directly into applications, systems, or devices, enabling real-time processing and decision-making without relying on external servers or cloud-based resources [44]. This technique is particularly useful in scenarios where low latency, high privacy, or offline capabilities are required [45].

Embedded models are typically optimized for performance and efficiency, often through techniques like model quantization, pruning, or distillation, to reduce their computational footprint. For instance, an embedded AI model might be used in a mobile app to perform natural language processing tasks, such as voice recognition or text prediction, directly on the device [46]. This approach ensures that the application can function smoothly and respond quickly, even in environments with limited or no internet connectivity.

Embedded models plays a key role in RAG systems (refer to section 2.2.3), where the model can index and retrieve relevant content directly from local or edge sources, enhancing real-time data access. In such systems, embedding content indexing within the device allows AI models to efficiently search through a predefined set of documents or knowledge bases, further reducing dependency on cloud-based retrieval services. This ability to perform RAG locally strengthens the system's response speed and privacy control, as data remains on the device.

In addition to mobile devices [46], embedded models are also increasingly used in edge computing, where data is processed closer to the source, such as in autonomous vehicles [47], industrial machines like Industrial Internet of Things [48], and smart home devices [49]. In these cases, the integration of RAG techniques into embedded models allows these systems to access indexed data on-site, whether for predictive maintenance or decision-making, ensuring that the right information is available when needed without cloud reliance. By embedding AI models within these systems, developers can achieve greater autonomy, faster response times, and enhanced security, as sensitive data does not need to be transmitted to external servers for processing.

**Fine-Tuned Models**
Fine-tuning is a technique used to adapt a pre-trained model to a specific task or dataset by further training it on a smaller, more focused dataset [50]. This approach is highly effective because it leverages the general knowledge acquired by the model during its initial training on large and diverse datasets, while refining its capabilities to excel in particular areas of interest.

For example, a general-purpose language model like GPT-4 can be fine-tuned for specific applications, such as sentiment analysis, legal document processing, or medical text generation. During fine-tuning, the model's parameters are adjusted to better align with the nuances of the new dataset, allowing it to produce more accurate and contextually relevant outputs for the specific task at hand. This process often involves training the model for fewer epochs and using a lower learning rate to prevent overfitting, ensuring that the model retains its broad knowledge base while improving its performance on the specialized task.

**Continued Pre-Training (CPT)**
The CPT approach involves extended pre-training on a domain or task, followed by supervised training on the downstream task [51]. Instead of training a language model from scratch, CPT allows reusing the existing model and incrementally enhancing its performance without starting over. Techniques like careful adjustment of learning rates and data scheduling help improve the model's capabilities without losing knowledge gained from earlier training stages [52]. CPT expands a model's general knowledge by further training on large, domain-specific data, improving its overall performance in a domain, while fine-tuning adapts a model for a specific task using a smaller, labeled dataset, optimizing it for that particular application.

For instance, if a consulting firm pre-trained a model on general architectural texts and building codes, they may later want it to analyze energy efficiency regulations for green building certifications. Initially, the model may struggle with specialized terminology. Through CPT, training on datasets like LEED guidelines and energy efficiency standards, the model can better understand sustainability-focused language. However, to specifically identify violations in energy compliance reports, the company would apply fine-tuning on a labeled dataset of compliance reports and violations, optimizing the model for this precise task without further broadening its overall domain knowledge.

### 2.5 Summary

The current landscape of AI models for image, video, and text generation is shaped by innovations from leading technology companies, leveraging advanced core technologies such as diffusion models and transformer models. Diffusion models are particularly effective in generating highly detailed and realistic images, refining visual outputs through iterative processes. In contrast, transformer models excel at understanding and generating sequential data, making them ideal for text and video generation tasks.

Many of the latest AI models combine these two technologies, capitalizing on their complementary strengths to enhance performance and versatility across various applications. These models are accessible through a range of platforms and methods. For instance, users can interact with these models in real-time through online playgrounds, integrate AI capabilities into existing applications via APIs, or deploy models locally using downloadable open-source projects.

Additionally, techniques such as fine-tuning and embedding are becoming increasingly important for customizing these models to specific tasks and environments. Fine-tuning adapts pre-trained models for specialized applications by refining them with targeted data, while embedded models allow AI capabilities to be integrated directly into applications and devices for real-time processing.

As AI continues to evolve, we can expect the development of even more sophisticated and accessible models, further expanding the possibilities for AI-driven creativity and functionality across various domains. The table below provides a detailed overview of the current state of these models, including their respective platforms and usage methods.

## 3 GENERATIVE AI FOR BUILDING RESEARCH AND INDUSTRY

This section of the manuscript delves into the application of generative AI in building research and industry, focusing specifically on two prevalent types of models: the transformer model (text-to-text) and the diffusion model (text-to-image), as shown in Figure 2. These models represent cutting-edge approaches in the field of artificial intelligence, offering unique advantages in processing and generating complex data sets pertinent to building design, construction, and management. By exploring the capabilities and distinctions of these models, we aim to elucidate how they can be effectively utilized to address various challenges in the building industry, enhancing both the efficiency and precision of predictive analytics and decision-making processes.

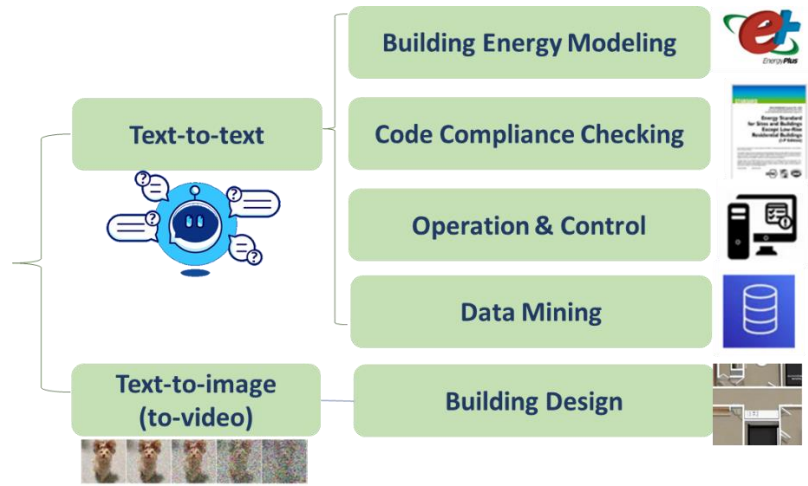

**Figure 2 Application for Gen AI in Building Industry and Research**

## 3.1 Text-to-text Generation
### 3.1.1 Code Compliance Checking

Using NLP for code compliance checking is not new. Lee et al. [53] presents a method to improve building code compliance checking by defining high-level, implementable methods that enhance low-level rule-checking procedures, thereby automating the translation of natural language building regulations into explicitly defined rules. The authors propose high-level computer programming operators designed to translate verb phrases in building act sentences into clear, actionable rules, ensuring minimal ambiguity and accurate representation of building object properties. They classified and filtered building-permit-related regulations from the Korean Building Act, extracting verb phrases to clarify the functions and arguments of the methods. These methods were organized into a four-level hierarchy based on targets, property types, checking types, and specific methods. The proposed methods translate verb phrases into building code sentences, retrieve data from a building model, and produce return values, which are then structured using IF-THEN-ELSE logic for rule translation. The intermediate code generated can be used in various rule-checking applications, supported by a reusable database for efficient compliance checking. The paper demonstrates the feasibility of this approach using KBimCode and developed rule-checking software, showing it to be more efficient in terms of time, effort, and cost compared to building low-level systems. This method also provides a universal standard for different building codes, potentially securing international consensus. By bridging the gap between natural language processing and building information modeling, this approach enables real-time application of regulations throughout the design phase, leading to reduced project durations and construction cost savings. The integration of new NLP methodologies, such as LLMs, into BIM-based code compliance checking systems is highlighted as a potential future direction.

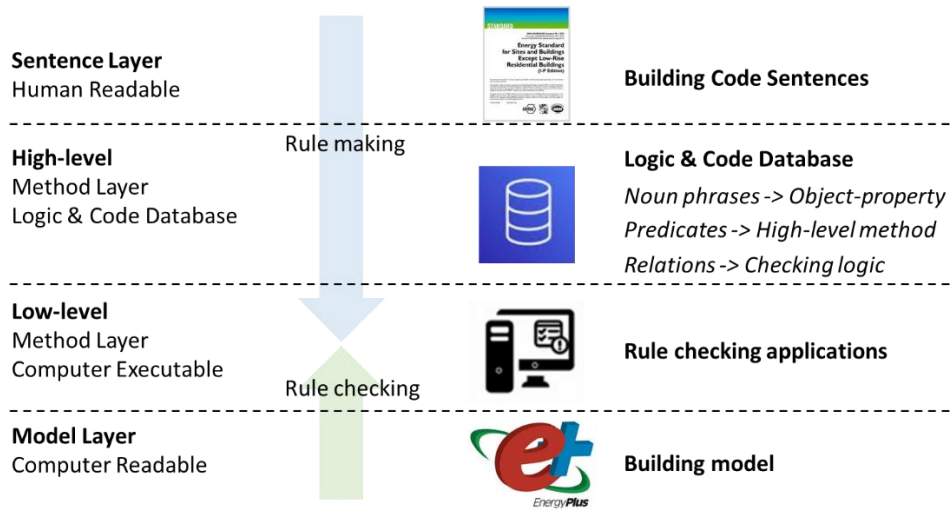

**Figure 3 Logic rule-based mechanism for compliance checking [53]**

Liu et al. [54] focused on developing an Automatic Compliance Checking (ACC) process for building design specifications by leveraging deep learning and LLMs. Their study involved evaluating GPT-based models through four distinct tests to determine their effectiveness in comprehending and processing regulatory documents. To enhance accuracy, the researchers fine-tuned their own model specifically trained on building specifications. The study aimed to compare different OpenAI models, highlighting the feasibility and potential of this approach in automating compliance checking within the Architecture, Engineering, and Construction (AEC) industry.

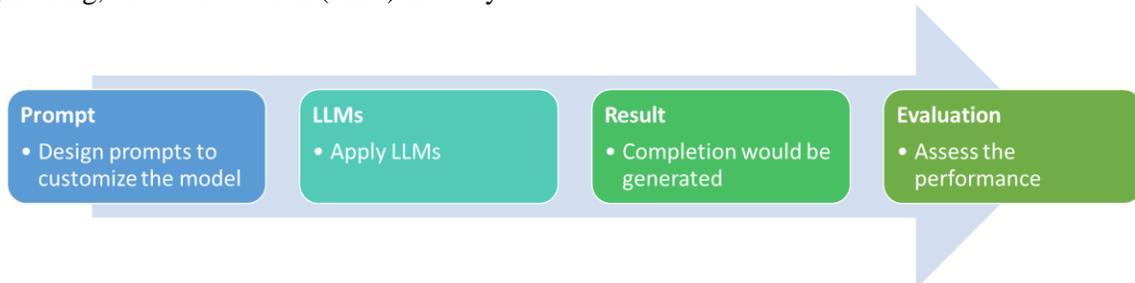

**Figure 4.    Design of Liu et al.'s Implementation [54]**

### 3.1.2 Control Support

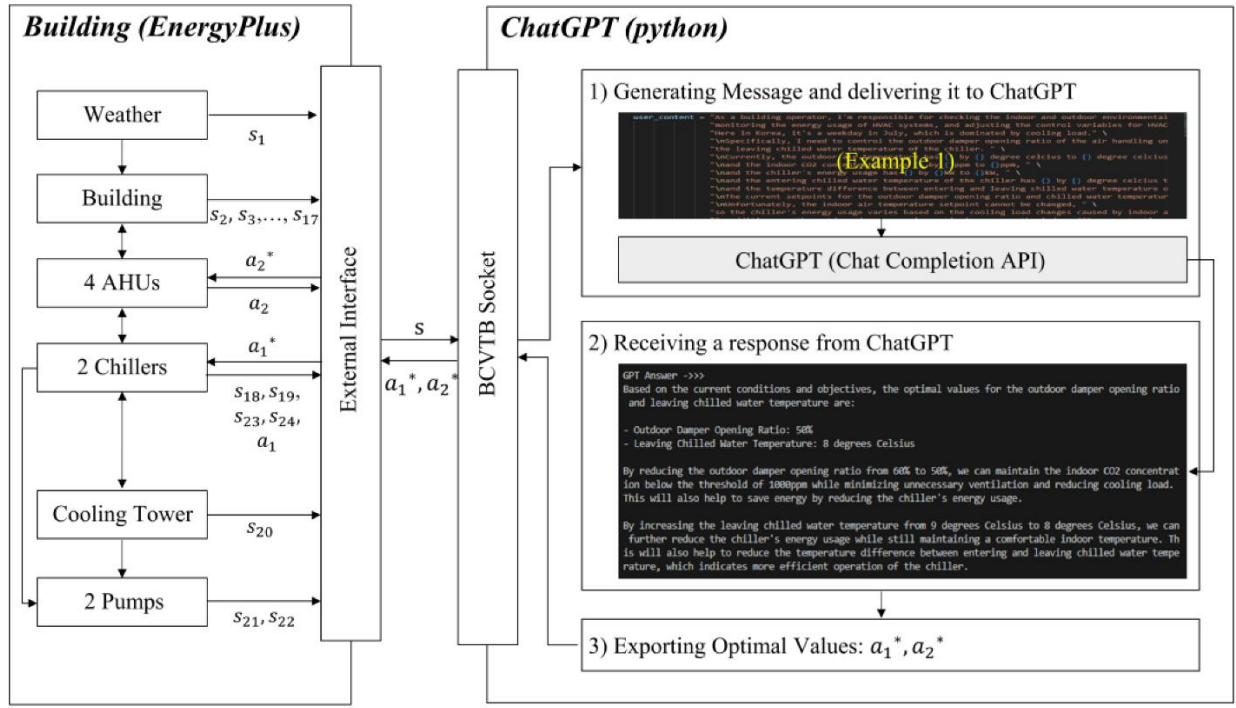

Figure 5.  Co-simulation of EnergyPlus and ChatGPT[55][§]

Ahn et al. [55] investigates the use of artificial intelligence, specifically ChatGPT and Deep Q-Network (DQN) models, to optimize the control of HVAC systems in a reference office building simulated using EnergyPlus. The goal was to minimize energy consumption while maintaining indoor CO2 levels below 1000 ppm. ChatGPT, utilizing its pre-trained language model capabilities via the OpenAI Chat Completion API, provided real-time decision-making without traditional training processes. In contrast, the DQN underwent 500 iterations over an 11-day period to refine its policy. Over a 3-day evaluation period, the ChatGPT control achieved a 16.8% reduction in energy use, whereas the DQN achieved a 24.1% reduction. Although the DQN showed higher energy savings, ChatGPT demonstrated significant potential by leveraging real-time building operational data and pre-learned domain knowledge. Future research will focus on fine-tuning ChatGPT for more specialized HVAC control tasks and managing complex building state information. The study highlights the potential of generative AI in autonomous building system operations, showcasing its capability to provide plausible control actions, such as adjusting outdoor damper openings and target leaving chilled water temperatures, to enhance energy efficiency.

### 3.1.3 Data mining

Zhang et al. [56] presents an automated data mining framework that combines maximal frequent itemset mining with generative pre-trained transformers (GPT) to identify energy waste patterns in building operational data. The framework improves the efficiency of extracting valuable operation patterns by reducing redundancy and transforming these patterns into prompts for GPT. This process automates the analysis of large datasets, liberating users from tedious manual evaluations. The framework was tested on a year's worth of data from a real-world building chiller plant system, successfully detecting various energy waste patterns, such as valve faults and improper device coordination, with a detection accuracy of 89.17% for energy waste patterns and 99.48% for normal operations. The study highlights the cost-effectiveness and competitive edge of using GPT for this purpose, with a total cost of $17.68 (compared to several weeks

---

of effort typically required by an engineer), demonstrating its potential to replace human analysts in identifying and addressing energy inefficiencies in building systems.

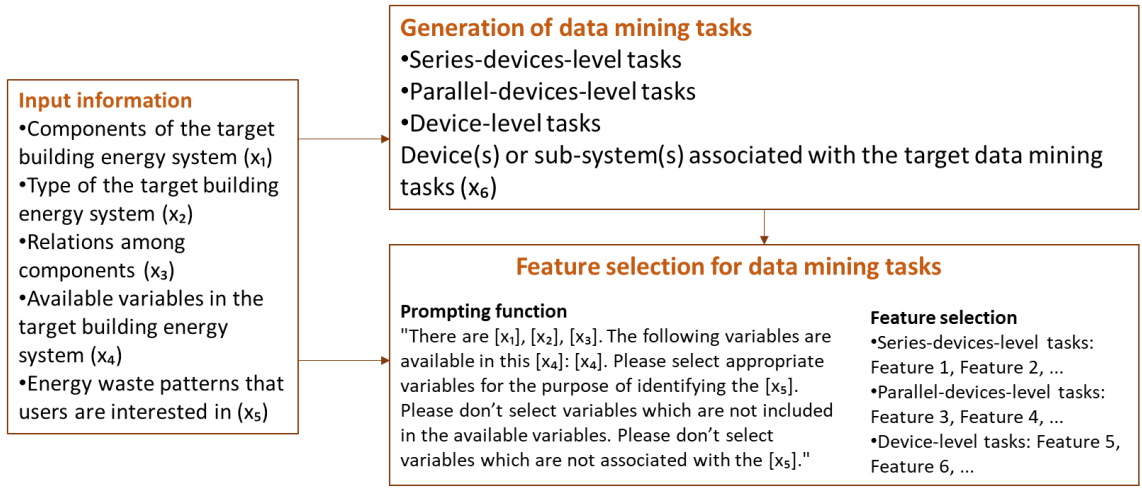

**Figure 6 Illustration of kernel density estimation-based transformation [56]**

### 3.1.4 Building Simulation Input Files Support

Zhang et al. [57] investigates the integration of LLMs, such as ChatGPT, with building energy modeling (BEM) software, specifically focusing on EnergyPlus [58]. It begins with a literature review highlighting the emerging trend of using LLMs in engineering modeling, though limited research exists on their application in BEM. The paper identifies several potential applications for LLMs in BEM, including generating or modifying IDF files, visualizing simulation outputs, conducting error analysis, co-simulation, extracting and training simulation knowledge, and optimizing simulations.

Three case studies are presented to demonstrate the transformative potential of LLMs in automating and optimizing BEM tasks, emphasizing the role of artificial intelligence in advancing sustainable building practices and energy efficiency. These studies highlight the importance of selecting appropriate LLM techniques, such as prompt engineering, RAG, and multi-agent systems, to enhance performance and reduce engineering efforts.

The paper also discusses the challenges associated with LLMs, including significant computational demands and potential consistency issues, but notes that ongoing technological advancements are addressing these limitations. The findings advocate for a multidisciplinary approach, combining expertise from AI and building modeling to effectively utilize LLMs in BEM. Future research should focus on creating specialized LLMs, such as "BEMGPT," tailored specifically for BEM to further sustainable building solutions.

Liang's another paper [9] explores the integration of LLMs like ChatGPT into building energy efficiency and decarbonization studies. It highlights the potential of LLMs to address various challenges in this field, including intelligent control systems, code generation, data infrastructure, knowledge extraction, and education. The paper emphasizes the transformative impact of LLMs in automating and optimizing tasks such as generating or modifying IDF files, visualizing simulation outputs, conducting error analysis, co-simulation, and simulation optimization. Case studies demonstrate the effectiveness of LLM techniques like prompt engineering, RAG, and multi-agent systems in enhancing BEM processes. Despite their promise, LLMs face challenges such as high computational costs, data privacy concerns, complexity in fine-tuning, and self-consistency issues. The paper calls for future research to improve LLMs for domain-specific tasks, develop multi-modal LLMs, and foster collaborative efforts between AI and energy experts to maximize the benefits of LLMs in building energy efficiency and decarbonization.

Jiang et al. [59] introduces Eplus-LLM, an innovative platform that leverages a fine-tuned LLM, to automate the process of BEM. The platform addresses the significant challenges in building design and analysis posed by the demanding modeling efforts, the need for expertise in simulation software, and the requisite building science knowledge. By utilizing a fine-tuned LLM, specifically T5 [60], Eplus-LLM can translate natural language descriptions of buildings into established models with various geometries, occupancy scenarios, and equipment loads. The system automates the generation of building models and simulation files, reducing over 95% of the manual modeling effort and achieving 100% accuracy, as validated by 152 test cases. The platform is robust to different tones, misspellings, omissions, and redundancies, making it highly adaptable and versatile.

Eplus-LLM operates by understanding human language through tokenization and embedding techniques, then generating building models and simulation results via the EnergyPlus simulation engine. This user-friendly human-AI interface significantly reduces the effort and dependency on specialized software for building modeling. The validation results demonstrated the platform's efficiency and accuracy, aligning with manual expert modeling while also showcasing robustness against various types of noise and unforeseen prompts.

Despite its success, the platform currently handles relatively simple modeling cases and is limited by objective conditions such as GPU availability, training time, and LLM performance. It struggles with complex geometries, multiple zones, and schedules, and cannot yet process interdependencies requiring more nuanced semantic understanding.

Future research directions include enhancing the platform's capability to handle more complex modeling scenarios, such as intricate zoning and detailed semantic descriptions, and further refining the LLM's performance through advanced prompt engineering and instruction techniques. These advancements aim to support large-scale building energy modeling and intelligent building management, thereby broadening the application of Generative AI in building design and development throughout the building lifecycle.

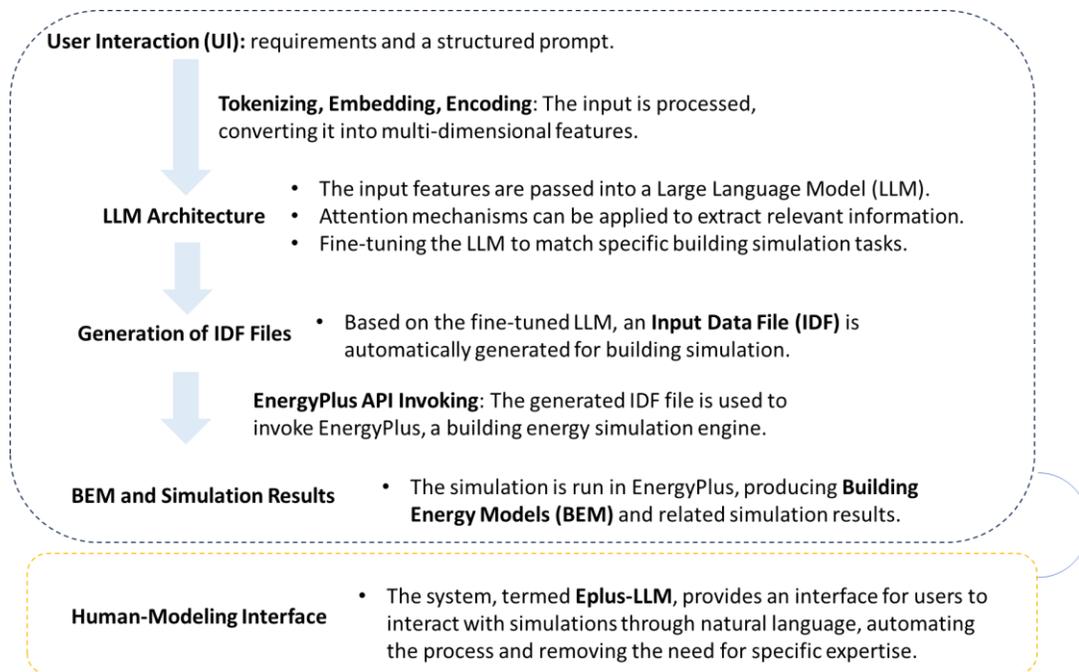

**Figure 7 The framework of the Eplus-LLM platform [59]**

## 3.2 Text-to-image Generation
### 3.2.1 Building style design

[Figure showing a table with columns "Trigger words (common prompts: building facade, storefront image)", "None", "Korea style", "Tokyo style", "Manhattan style", and a row of "Output images" containing generated building facade images for each style.]

**Figure 8.     Images generated by entering various scale-specific building façade design style prompts [61]**

Jo et al. [61] proposes a novel approach and tool, named AIBIM-Design, for generating photorealistic building façade images using Generative AI. The approach addresses the significant time and resource demands of existing tools in architectural visualization, which are essential throughout a project's lifecycle for communicating designs to stakeholders and facilitating design improvements. Leveraging generative

AI, the proposed method offers a more efficient and scalable solution, capable of producing a broader range of design alternatives. The feasibility of this approach is demonstrated through intensive testing within a specific design context, focusing on generating façade design alternatives that reflect local identities. The study confirmed that including region-specific prompts in the AI model leads to higher quality and more detailed images. To achieve this, the model was trained using a dataset paired with text data, obtained and preprocessed from street-view images of different regions. The trained model was validated with neighborhood facilities in Seoul, successfully transforming mass geometry into photorealistic images.

The research resulted in the generation of approximately 2000 façade images, consuming significant but manageable hardware resources. The AIBIM-Design tool, developed from this study, is accessible to both architectural experts and the public, allowing the generation of real-life images that represent diverse design identities from mass-type models. Future directions include expanding the training models horizontally to cover various building façade types and categories, and vertically to incorporate different kinds of visualizations beyond façades. This generative AI-based visualization tool demonstrates the potential to enhance financial and temporal efficiency in the design communication phase of architectural projects, providing a user-friendly interface for generating design alternatives that resonate with the local identity of the area.

### 3.2.2 Floor plan generation

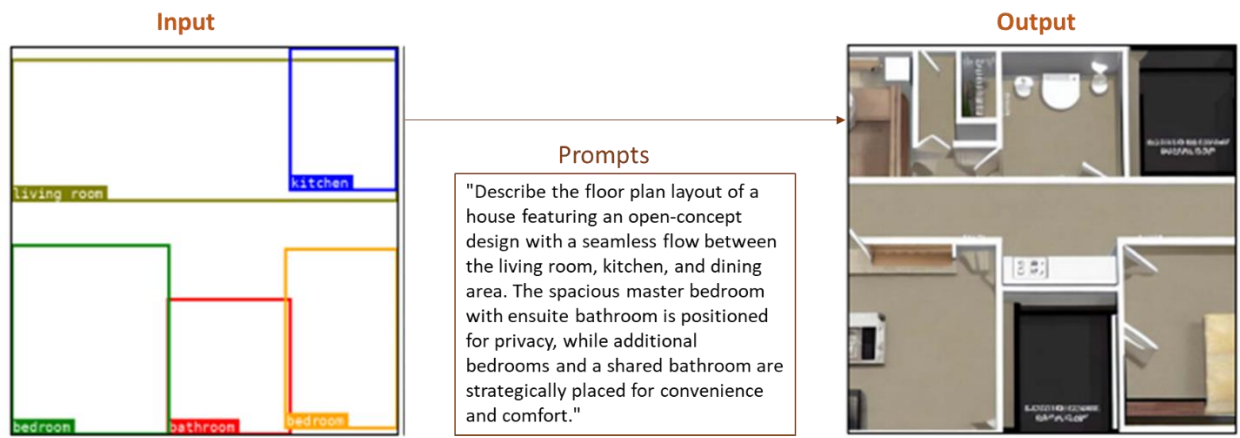

**Figure 9.** Existing generative models can generate layout diagrams of rooms based on input text and can also be controlled accordingly based on input layouts [7]

Li et al. [7] explores the application of generative AI techniques in the generation of architectural floor plans, emphasizing advancements that allow architects to incorporate more conditional constraints into the design process. Architects can provide text data, such as client design requirements and architectural standards, and image data, including site plans and floor layout diagrams, to generative models. These inputs enable models to produce detailed and contextually relevant floor plans. The paper highlights the use of "Scene Graphs," which describe the elements and their interrelations within a scene, making them suitable for depicting architectural floor plans. Technologies like SceneGenie integrate diffusion models to accurately generate these plans using Scene Graphs. Moreover, advanced generative models such as Stable Diffusion and Imagen can refine the generation process through text prompts and layout controls, producing comprehensive architectural designs based on input text. Despite these advancements, the generated images often lack professional standards and rational layout adherence. However, the incorporation of additional constraints, like bounding boxes, shows promise for better alignment with architectural design considerations. The paper concludes that integrating professional architectural data with computational data, as illustrated by layout and segmentation masks, enhances the generative process, making it more efficient and capable of producing high-quality architectural floor plans.

## 4 CASE STUDIES: BUILDING ENERGY CODE COMPLANCE CHECKING

### 4.1 Demo Development

The authors leverage a LLM for building code compliance checking as a case study, with the primary purpose of using LLMs for content indexing and RAG. This approach enables the extraction of detailed requirements from building energy code documents such as ASHRAE 90.1 [62] and International Energy Conservation Code (IECC) [63]. For example, it can precisely extract specific R-value requirements for a particular climate zone and code version, such as ASHRAE 90.1 2019. The system employs Chat-GPT-4 and embedded models provided by the Azure OpenAI [64] API, with no fine-tuning applied, to achieve these tasks effectively, as shown by Table 3.

**Table 3 Model Applied**

|  | Model Name | Version |
|---|---|---|
| Embedded | Text-embedding-ada-002 | 2 |
| GPT | Gpt-4 | 1106-Preview |
| Platform | Azure OpenAI Service | westus |

Figure 10 illustrates our customized RAG process designed specifically for building energy code compliance checking. The process begins with input parsing, where various documents such as building plans, energy models, and code specifications are extracted and processed. These inputs are then transformed into vector representations using an embedded model and stored in a vector database, which supports efficient retrieval of relevant code sections. An initial prompt is generated based on the parsed inputs and retrieved data, which is then processed by a LLM via an API to produce detailed responses related to code compliance. A caching mechanism is employed to store frequently accessed results, enhancing the efficiency of the process. The generated outputs are filtered to ensure relevance and accuracy before being presented through the application's user interface, allowing users such as building inspectors and designers to review the compliance results and take necessary actions. This tailored RAG approach ensures thorough and efficient evaluation of building designs against energy codes.

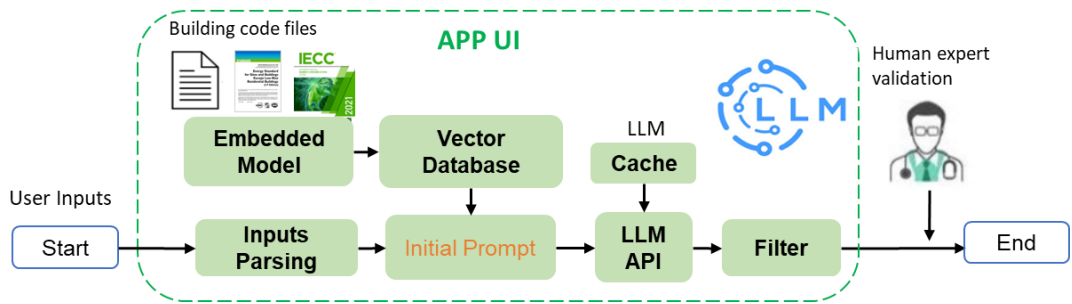

**Figure 10 Flowchart for Customized RAG for Building Energy Code Compliance Checking**

We also developed a User Interface (Chatbot) as shown by Figure 11.

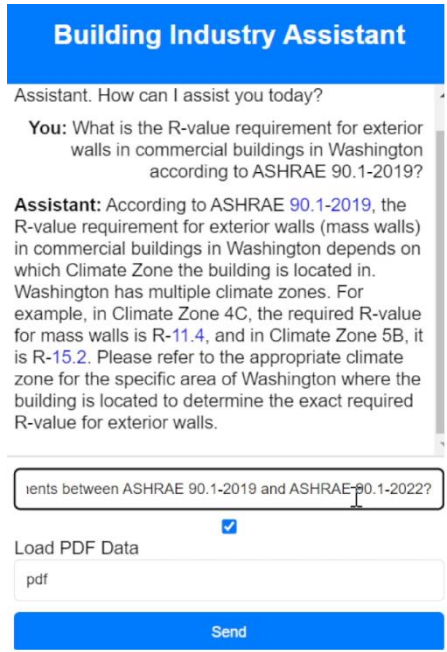

**Figure 11 Demo Chabot for Building Energy Code Compliance Checking**

**Table 4 Sample Prompts Testing**

| Loaded Code | Input Prompts | Outputs |
|---|---|---|
| ASHRAE Standard 90.1 | "What is the R value requirement for Washington State for external wall for commercial buildings?" | "Washington State primarily falls under Climate Zone 4C, and according to ASHRAE Standard 90.1, the requirements are as follows:<br>**Mass Walls**:<br>Minimum R-value: R-9.5 continuous insulation (c.i.)<br>Maximum U-value: U-0.104<br>**Metal Building Walls**:<br>Minimum R-value: R-0 + R-15.8 c.i.<br>Maximum U-value: U-0.060<br>**Steel-Framed Walls**:<br>Minimum R-value: R-13 + R-7.5 c.i.<br>Maximum U-value: U-0.064<br>**Wood-Framed and Other Walls**:<br>Minimum R-value: R-13 + R-3.8 c.i. or R-20<br>Maximum U-value: U-0.064." |
| ASHRAE Standard 90.1 | "What are the requirements for economizers in air conditioning systems?" | "**Air Economizer Design Capacity:**<br>Air economizer systems must be capable of modulating outdoor air and return air dampers to provide up to 100% of the design supply air quantity as outdoor air for cooling.<br>**Control Signal**:<br>Economizer controls must be capable of and configured to sequence the dampers with the mechanical cooling equipment. The control should not rely solely on mixed-air temperature, except in systems controlled from space temperature, such as single-zone systems.<br>**High-Limit Shutoff:** |

|  |  |  | All air economizers must automatically reduce outdoor air intake to the design minimum when outdoor air intake will no longer reduce cooling energy use. The specific high-limit shutoff control types and set points vary by climate zone and are detailed in Table 6.5.1.1.3.<br>**Dampers**:<br>Exhaust/relief and outdoor air dampers must meet specific sealing requirements, and return dampers must meet the requirements for motorized exhaust/relief dampers.<br>**Relief of Excess Outdoor Air**:<br>Systems must provide a means to relieve excess outdoor air during air economizer operation to prevent overpressurizing the building. The relief air outlet should be located to avoid recirculation into the building." |
|---|---|---|---|
|  | ASHRAE Standard 90.1 | "What are the guidelines for solar-ready roofs?" | **Solar-Ready Zone**:<br>A designated solar-ready zone must be allocated on the roof or other suitable area of the building to accommodate future solar photovoltaic or solar thermal installations. This zone should be free of obstructions such as HVAC equipment, vents, or other fixtures.<br>**Solar-Ready Area**:<br>The solar-ready zone must meet specific size requirements, generally based on a percentage of the total roof area or a minimum required area. The exact area may vary depending on the building's location, type, and other factors.<br>**Pathways for Conduits**:<br>Pathways must be provided for routing conduit or piping from the solar-ready zone to the electrical service panel or hot water system. These pathways should be planned to minimize the length and complexity of future installations.<br>**Structural Load**:<br>The roof structure should be designed to support the additional load of solar panels, including both dead loads and live loads. Structural engineers must consider these loads in the initial design phase to avoid future complications.<br>**Documentation**:<br>The construction documents must include details of the solar-ready zone, conduit pathways, and any structural modifications required to accommodate solar installations. This information should be easily accessible for future reference by building owners or contractors." |

To test our demo chatbot for building energy code compliance checking, we developed a series of scenarios focused on the solar-ready provisions as outlined in the building energy codes, as shown in Table 4Table 1. The chatbot was assessed on its ability to accurately identify and interpret key requirements, such as the allocation of a solar-ready zone on the roof, ensuring the zone is free from obstructions like HVAC equipment or vents. We also evaluated the chatbot's capacity to verify that the solar-ready area meets specific size requirements based on the building's characteristics and location. Additionally, the chatbot was tested on its ability to confirm that proper pathways for conduits are planned, ensuring minimal complexity for future solar installations. Furthermore, the chatbot was required to check if the structural load calculations were included in the design to support the additional weight of solar panels. Lastly, we tested the chatbot's ability to ensure that all relevant documentation, including details of the solar-ready zone and any structural modifications, was included in the construction documents for future reference. This comprehensive testing approach ensured that our chatbot could effectively assist in building energy code compliance checking with a focus on solar-ready requirements.

### 4.2 Future works

As we look to the future, several key areas for further development and enhancement of our demo chatbot have been identified. One significant avenue is the integration of advanced image reading capabilities, including Optical Character Recognition (OCR) for extracting text from images. While basic OCR for recognizing words in images is relatively straightforward, the challenge lies in interpreting complex visual information. For instance, in a floor plan, automatically identifying and distinguishing different zones, locating features such as balconies, or understanding the spatial relationships between elements requires sophisticated image analysis and pattern recognition techniques. This is an area that will need more research and development to ensure accurate and reliable interpretations of such details.

Another exciting potential direction is the expansion of our chatbot beyond energy code compliance to serve as a comprehensive code compliance assistant, or "code copilot." This could include not only building energy codes but also fire codes, environmental regulations, construction standards, and other relevant codes. By broadening the scope, our chatbot could provide a more holistic solution for the AEC industry, assisting professionals in ensuring that all aspects of their projects are compliant with various regulations.

Furthermore, we envision enhancing the chatbot's capabilities to read and interpret tables, which are often used in code documents to convey detailed requirements and specifications. The ability to accurately extract and understand information from tables would significantly improve the chatbot's utility in code compliance checking.

In summary, the future development of our demo chatbot will focus on advanced image and table reading, expanding its application to various types of codes, and evolving it into a more intelligent and versatile "code copilot" that can support a wide range of compliance needs within the AEC industry. These developments will not only increase the functionality of the chatbot but also make it a valuable tool for professionals seeking to navigate the complexities of regulatory compliance.

## 5 DISCUSSIONS

### 5.1 Need for current building research and industry
#### 5.1.1 Image Processing Techniques

The building industry is increasingly integrating digital technologies to improve efficiency, accuracy, and compliance with regulatory standards. One of the critical areas where advancements are needed is in image processing techniques. While current OCR technologies can extract text from images effectively, the challenge remains in interpreting complex visual data, such as floor plans, blueprints, and site layouts. For instance, automatically identifying and categorizing different zones, detecting specific architectural features like balconies, or understanding the spatial relationships between elements on a floor plan requires more sophisticated image analysis algorithms. These tasks are crucial for automating compliance checks, design reviews, and construction monitoring, yet they pose significant challenges due to the complexity and variability of building designs. Future research must focus on developing advanced image processing tools that can reliably interpret these details, enhancing the automation capabilities in the building industry and reducing the reliance on manual inspections.

### 5.1.2 Shape of the products

Another critical aspect of the current building industry's needs is the ability to accurately assess and manage the shapes and dimensions of construction products. The geometric complexity of modern building components, from prefabricated panels to intricate façade designs, necessitates advanced modeling and analysis tools. Traditional methods of measuring and validating product shapes are often labor-intensive and prone to errors. Therefore, integrating AI-driven shape recognition and analysis technologies into the construction workflow could significantly improve the precision and efficiency of product design, manufacturing, and quality control. These technologies can also assist in ensuring that products meet specific regulatory standards and fit seamlessly within the overall building design, thereby reducing the risk of costly rework and project delays.

### 5.1.3 AI workforce development

As AI techniques grow, people start to suggest that revising architecture education to include AI-related courses is crucial for training future architects who can harness the latest AI tools to design energy-efficient, resilient buildings [65]. However, the development of architecture courses must evolve to encompass the latest advancements in AI, particularly focusing on both diffusion and transformer-based technologies. Diffusion models, which excel in image generation, have already gained traction in fields such as animation and PC game design, and their potential in architecture, particularly for building design, is immense. These models can generate realistic renderings, conceptual designs, and even assist in visualizing complex architectural forms with fine detail. In contrast, transformer models offer significant benefits to building engineers and modelers who rely on NLP techniques. These professionals can utilize AI to automate building code compliance checks, optimize energy performance, and generate detailed documentation. While diffusion models and transformers represent two distinct technological tracks, they offer complementary capabilities in the architecture field: diffusion models enhancing the visual and creative design process, and transformer models empowering engineers and modelers through text-based insights, analysis, and automation. Therefore, architecture curricula must integrate both tracks, equipping future professionals with the skills needed to leverage these AI advancements in diverse aspects of building design, engineering, and operation.

## 5.2 Possible Other Application Areas
### 5.2.1 Technical Assistance

Beyond the immediate needs of the building industry, AI technologies have the potential to revolutionize technical assistance and workforce training across various fields. Leveraging our lab's advanced capabilities, we can harness AI to create innovative training tools that are not only cost-effective but also provide safer alternatives to traditional methods. For instance, by employing diffusion models, we can generate high-quality educational videos with integrated audio instructions, offering a comprehensive and immersive learning experience. These AI-generated videos could be particularly useful in training scenarios where real-world simulations would be prohibitively expensive or dangerous, such as refrigerant charge procedures and the associated risks of leaks or fire hazards. Such AI-driven simulations allow trainees to safely engage with realistic scenarios, ensuring they are well-prepared for actual fieldwork.

### 5.2.2 Building Industry Code Copilot

In addition to technical assistance, AI can play a transformative role in supporting the building industry through tools like a "code copilot." This concept involves developing AI-powered assistants that can help professionals navigate and comply with complex building codes, such as energy codes, fire safety regulations, and environmental standards. For example, integrating AI with platforms like the Modelica Building Library could automate the process of ensuring that building designs meet regulatory requirements. The AI could automatically interpret and apply relevant codes to specific design elements, streamlining the compliance process and reducing the likelihood of human error. This would be especially valuable in large-scale projects where manual code compliance checks can be time-consuming and prone to oversight.

### 5.2.3 Control Diagram Diagnostic Tool

Another promising application of AI in the building industry is in the development of control diagram diagnostic tools. These tools would analyze control diagrams—essentially the blueprints for building automation and HVAC systems—and identify potential issues or inefficiencies. By applying machine learning algorithms to the visual and functional aspects of these diagrams, AI could automatically detect anomalies, suggest improvements, and ensure that the systems are optimized for energy efficiency and performance. This would be particularly useful during the design phase, where early detection of control system issues can prevent costly revisions later in the project lifecycle.

In summary, the potential applications of AI extend far beyond the immediate needs of the building industry. From advanced technical training tools to AI-driven code compliance assistants and diagnostic tools for control systems, the integration of AI across these areas promises to enhance efficiency, safety, and compliance in various technical fields. These advancements will not only streamline processes but also empower professionals with the tools they need to meet the challenges of modern construction and building management.

## 6 CONCLUSIONS

The integration of generative AI into the building industry holds substantial promise for transforming various aspects of design, compliance, and training. This review has underscored the potential of generative AI technologies, such as LLMs, to automate and enhance processes that were previously labor-intensive and prone to human error. By leveraging the capabilities of LLMs in tasks like energy code compliance checking, building design optimization, and workforce training, the building industry can achieve significant improvements in efficiency, accuracy, and safety.

Generative AI, particularly in the form of LLMs, offers a powerful tool for streamlining energy code compliance. By automating the analysis of architectural plans against complex regulatory frameworks, these models can rapidly identify areas of non-compliance, significantly reducing the time and effort required for manual reviews. This capability is crucial in addressing the increasing complexity of building codes and the growing demand for sustainable building practices.

Moreover, the application of generative AI extends beyond compliance checking. The ability of AI to generate realistic simulations and design alternatives presents new opportunities in architectural design and training. For instance, AI-generated videos and simulations provide cost-effective and safe training solutions for complex scenarios, such as handling hazardous materials or optimizing HVAC systems. These applications not only improve training outcomes but also ensure that the workforce is better prepared for real-world challenges.

As AI technologies continue to evolve, their role in the building industry is expected to expand. Future developments may include advanced image processing techniques for interpreting complex visual data in architectural plans, as well as the creation of AI-driven tools for comprehensive code compliance across various regulatory domains, including fire safety, environmental regulations, and construction standards. Additionally, the potential for AI to serve as a "code copilot" could revolutionize the way professionals navigate and apply building codes, further enhancing the efficiency and accuracy of design and construction processes.

In conclusion, generative AI represents a transformative force in the building industry, offering new pathways for innovation in design, compliance, and training. By embracing these technologies, the industry can achieve greater sustainability, safety, and operational efficiency, setting the stage for continued advancements in building practices and regulatory adherence. The future of the building industry will undoubtedly be shaped by the integration of AI, paving the way for smarter, more responsive, and more sustainable construction practices.

## AUTHORSHIP CONTRIBUTION STATEMENT

**Hanlong Wan**: Conceptualization, Methodology, Validation, Formal analysis, Software, Writing - Original Draft, Visualization, Investigation. **Jian Zhang**: Supervision, Funding acquisition, Writing - Review & Editing. Project administration. **Yan Chen**: Conceptualization - AI for FDD, Writing - AI in Building Control, Writing - Review & Editing. **Weili Xu**: Conceptualization - Commercial Code Compliance Checking, Writing - Code Copilot Application, Writing - Review & Editing. **Fan Feng**: Visualization – Outline Figure.

## DECLARATION OF COMPETING INTEREST

The authors declare that they have no known competing financial interests or personal relationships that could have appeared to influence the work reported in this paper.

## ACKNOWLEDGMENTS

The authors would like to express their gratitude for the support of the U.S. Department of Energy's Office of Energy Efficiency and Renewable Energy (EERE) through Battelle Memorial Institute under Contract No. DE-AC05-76RL01830. The authors also acknowledge Lingzhe Wang for his assistance with formatting and technical editing.